\documentclass[10pt,twocolumn,letterpaper]{article}
\usepackage[accsupp]{axessibility}
\usepackage{iccv}
\usepackage{times}
\usepackage{epsfig}
\usepackage{graphicx}
\usepackage{amsmath}
\usepackage{amssymb}
\usepackage{booktabs}
\usepackage{multirow}
\usepackage{bm}
\usepackage[normalem]{ulem}
\usepackage{color}
\usepackage{diagbox}
\usepackage{multirow}
\usepackage{cellspace}
\usepackage{subfigure}
\usepackage{cite}
\usepackage[table,xcdraw]{xcolor}
\definecolor{purple}{RGB}{128,0,128}

\usepackage[pagebackref=true,breaklinks=true,letterpaper=true,colorlinks,bookmarks=false]{hyperref}
\usepackage[capitalize]{cleveref}
\crefname{section}{Sec.}{Secs.}
\Crefname{section}{Section}{Sections}
\Crefname{table}{Table}{Tables}
\crefname{table}{Tab.}{Tabs.}
\iccvfinalcopy 

\def\iccvPaperID{5894} 

\ificcvfinal\fi

\begin{document}

\title{When Prompt-based Incremental Learning Does Not Meet Strong Pretraining}

\author{Yu-Ming Tang$^{1,3}$
\qquad Yi-Xing Peng$^{1,3}$
\qquad Wei-Shi Zheng$^{1,2,3}$\thanks{Corresponding author}\\\\
{$^1$School of Computer Science and Engineering, Sun Yat-sen University, China}\\
{$^2$Peng Cheng Laboratory, Shenzhen, China}\\
{$^3$Key Laboratory of Machine Intelligence and Advanced Computing, Ministry of Education, China}\\
{\tt\small \{tangym9, pengyx23\}@mail2.sysu.edu.cn}
\qquad {\tt\small wszheng@ieee.org}}

\maketitle
\ificcvfinal\thispagestyle{empty}\fi

\begin{abstract}
    Incremental learning aims to overcome catastrophic forgetting when learning deep networks from sequential tasks.
    With impressive learning efficiency and performance, prompt-based methods adopt a fixed backbone to sequential tasks by learning task-specific prompts.
    However, existing prompt-based methods heavily rely on \textbf{strong} pretraining (typically trained on ImageNet-21k), and we find that their models could be trapped if the potential gap between the pretraining task and unknown future tasks is large.
    In this work, we develop a learnable \textbf{A}daptive \textbf{P}rompt \textbf{G}enerator (APG).
    The key is to unify the prompt retrieval and prompt learning processes into a learnable prompt generator.
    Hence, the whole prompting process can be optimized  to reduce the negative effects of the gap between tasks effectively. 
    To make our APG avoid learning ineffective knowledge, we maintain a knowledge pool to regularize APG with the feature distribution of each class. 
    Extensive experiments show that our method significantly outperforms advanced methods in exemplar-free incremental learning without (strong) pretraining.
    Besides, under strong pretraining, our method also has comparable performance to existing prompt-based models, showing that our method can still benefit from pretraining. 
    Codes can be found at \url{https://github.com/TOM-tym/APG}
    
\end{abstract}

\section{Introduction}
\label{sec:intro}
\begin{figure}
    \centering
      \includegraphics[width=\linewidth]{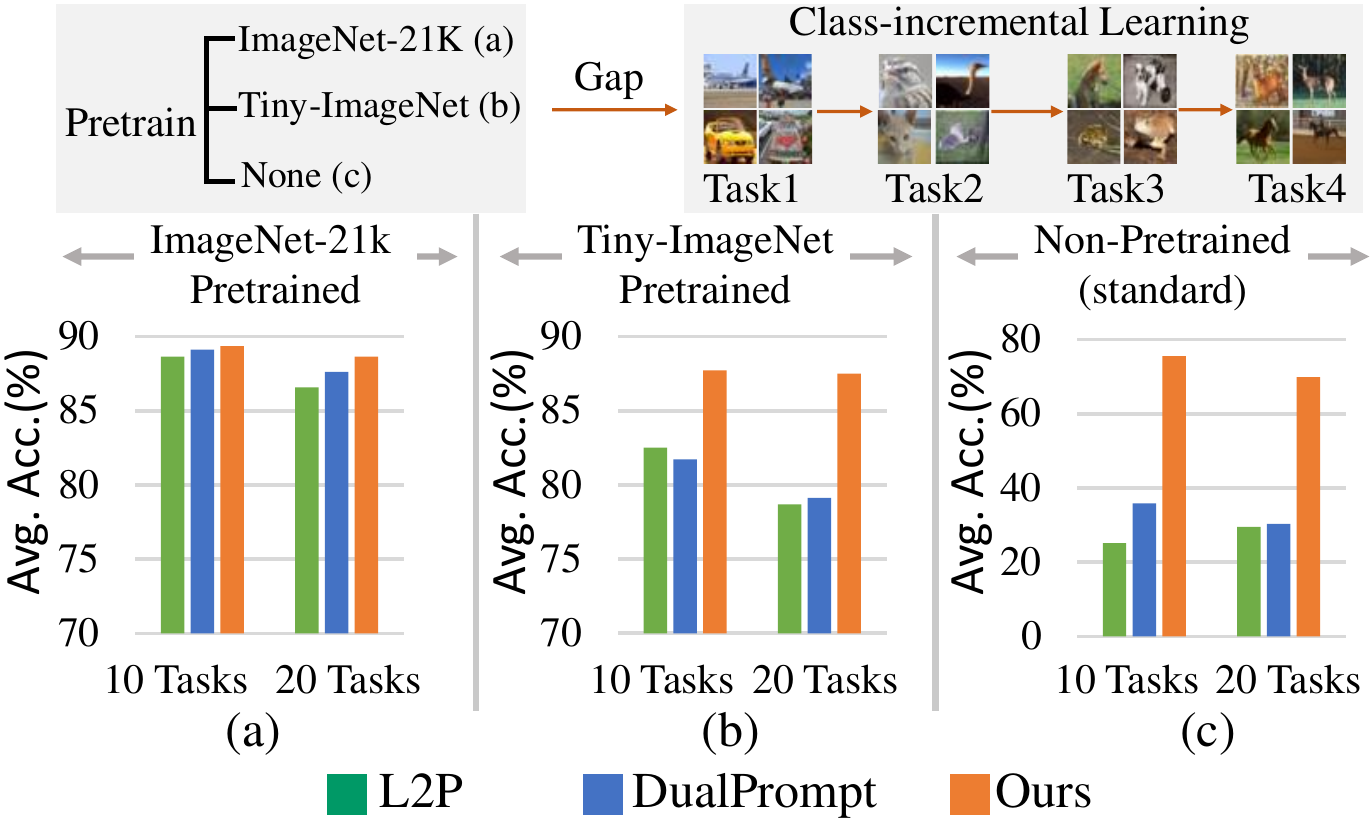}
      \caption{
      Experimental comparison between our adaptive prompting scheme and other prompted-based methods~\cite{l2p, dualprompt} on the CIFAR100~\cite{cifar100} dataset. 
      `Non-pretrained' (a standard protocol in incremental learning) means the data from the first task is used to pretrain the backbone.
      (a) With an intensive pretrained backbone, all three methods perform well. 
      (b) When swapping to Tiny-ImageNet (200 classes) pretrained weights, the performance of other methods clearly drops, while ours does not.
      (c) Our method significantly outperforms other methods in the presence of a large semantic gap between pretraining task and unknown future tasks.
      }
    \label{fig:intro}
  \end{figure}

Deep neural networks (DNNs) have become powerful tools in various fields~\cite{resnet, vit, deit, gao2022adamixer, li2022dn, i3d}.
However, when facing sequential training tasks, DNNs learn new tasks along with severe performance degradation on previous tasks in the absence of old data, which is the notorious catastrophic forgetting~\cite{forget1, forget2, forget3}.
Incremental learning aims to overcome catastrophic forgetting in DNNs, pushing DNNs toward complex real-world applications, \eg AI robotics~\cite{fang2020graspnet,sundermeyer2021contact,breyer2020volumetric,mousavian20196} or self-driving~\cite{held2016learning, prianto2020path, gupta2021deep, mutz2017following}.
Previous works usually maintain a memory buffer with a handful of old samples for rehearsal when learning new tasks~\cite{icarl, der, ucir, bic, dmc, inthewild, Imagine, foster}.
Since keeping old data may be infeasible due to privacy/storage concerns, another branch of work~\cite{SDC, PASS, SSRE} explores exemplar-free incremental learning, which tunes the network based on the introduced priors but the performance still far falls behind those of rehearsal-based methods.

Recently, an appealing development\cite{sprompt, l2p, dualprompt} based on prompting\cite{vpt, promptNLP_1, promptNLP_2, promptNLP_3, P_tuning_v2} manages to encode knowledge into sets of prompts to steer a frozen backbone for handling sequential tasks. In addition to impressive performance, it has several benefits.
(1) The catastrophic forgetting problem is effectively alleviated since the backbone is fixed; (2) learning prompts instead of backbone significantly reduces training costs and improves learning efficiency; (3) prompt-based methods are free from keeping exemplars.
To employ prompts for task-agnostic class-incremental learning, a crucial step is to select task-specific prompts given any input images.
Existing methods maintain a prompt pool and retrieve prompts by directly computing the similarity between the image feature extracted by the pretrained model and the prompts in the pool, which is simple yet effective with a strong pretrained model.
However, as the pretrained model dominates the retrieval, such a non-learnable retrieval process will be problematic because the future tasks are unknown and the gap between the pretraining task and the unknown future tasks could be large. 
As in \cref{fig:intro}~(c), when the first task in incremental learning is used for pretraining, the classes in the pretraining task are totally different from other tasks, which we refer to as a semantic gap that degenerates existing models.
Although the semantic gap between domains is also studied in some works such as transfer learning\cite{tan2018survey, pan2010survey, he2019rethinking}, their works do not consider the forgetting problem in sequential tasks.
It is necessary to emphasize that the intention of this work is \textbf{NOT} to refuse pretraining but to propose a more general method that does not rely heavily on strong pretraining and can benefit from it if task-related pretraining is available. For more experiments regarding the necessity of our work, please refer to the supplementary materials.

In this work, we develop a learnable \textbf{A}daptive \textbf{P}rompt \textbf{G}enerator (APG) to effectively bridge the potential gap between pretraining tasks and unknown future tasks.
The core of our method is to unify the prompt retrieval and the prompt learning process into a learnable prompt generator.
In this way, the whole prompting process can be optimized to reduce the negative effects of the gap between tasks effectively. 
Besides, rather than retrieving prompts from a fixed-size prompt pool, learning to generate prompts enhances the expression ability of prompts.
As a result, the APG can be applied to a model without strong pretraining, and notably, the employment of APG does not discount the effort on overcoming forgetting since the backbone is still fixed. 

For incremental learning, APG holds an extendable prompt candidate list for aggregating knowledge from seen tasks into a group of prompts.
To adaptively prompt the backbone, the knowledge aggregation in APG is conditioned on the immediate feature from the backbone.
In addition, we form a knowledge pool to summarize the knowledge encoded in the feature space.
The summarized knowledge is further used to regularize the AGP to prevent it from learning ineffective knowledge.

In summary, our contributions are as follows. 
(1) We propose a learnable adaptive prompt generator (APG) to reduce the negative effects of the gap between the pretraining task and unknown future tasks, which is critical but ignored by previous work.
Our adaptive prompting eases the reliance on intensive pretraining. 
(2) To regularize APG, we propose the knowledge pool, which retains the knowledge effectively with only the statistics of each class.
(3) The extensive experiments show that our method significantly outperforms advanced exemplar-free incremental learning methods without pretraining. Besides, under strong pretraining, our method also achieves comparable satisfactory performance to existing prompt-based models.

\section{Related Work}
\noindent\textbf{Non-pretrained Class-incremental Learning.}
Rehearsal-based class incremental learning methods~\cite{podnet, ucir, DDE, afc, CSCCT, GCR, Imagine, Mnemonics, rainbow} have access to a handful of old samples.
With the retained samples, several works propose to distill old knowledge to the current network~\cite{ucir, podnet, DDE, CSCCT, Imagine} or to maintain the old feature space~\cite{afc, GCR}.
Different from rehearsal-based methods, we propose to get rid of the memory buffer and bridge the gap between old and new tasks with a learnable module.
A straightforward idea to handle incremental learning is to dynamically expand the network for each task~\cite{serra2018overcoming, mallya2018packnet, mallya2018piggyback, ke2020continual,der, itaml, foster,ccgn,random_path}.
Although these methods are more intuitive, their methods often require careful design of network architecture and often require a memory bank like rehearsal-based methods.
There also exists exemplar-free methods aiming to learn sequentially tasks without saving any images.
These works propose to estimate the semantic drift~\cite{SDC}, or to keep knowledge in class prototypes~\cite{PASS, SSRE}. 
Although exemplar-free methods proposed an appealing prospect for incremental learning, they typically introduce manually designed priors to the learning of new tasks, and human-designed priors are less generalizable, which makes their results unsatisfactory.

\vspace{3pt}
\noindent\textbf{Prompt-based Class-incremental Learning.}
Inspired by prompting in natural language processing, some methods investigate prompt-based incremental learning and achieve great success~\cite{sprompt, l2p, dualprompt}. 
These methods are generally inherited directly from NLP, utilizing a pretrained transformer as a base.
S-prompt~\cite{sprompt} focuses on domain-incremental learning and learns different prompts across domains.
For class incremental learning, L2P~\cite{l2p} and DualPrompt~\cite{dualprompt} first propose to learn a pool of prompts and 
query the prompts based on the feature extracted by the pretrained backbone.
DualPrompt~\cite{dualprompt} further proposes to attach complementary prompts to divide old knowledge into the general one and the expert one.
The main idea of prompt-based methods is to encode knowledge from old tasks into sets of vectors (\ie prompts) and retrieve them to instruct the backbone when the old data is not accessible.
These methods are in need of a strong pretrained backbone to assist such a retrieval process.
Different from their methods, we propose a unified prompting scheme that eases the reliance on intensive pretraining and makes it suitable for both pretrained and non-pretrained scenarios.

\section{Adaptive-prompting for Incremental Learning}
\label{sec:method}
\subsection{Preliminary}
\label{sec:pre}
\noindent\textbf{Class-incremental learning.} \ \ 
For a deep model $\Phi(f(\cdot))$ consisting of a classifier $\Phi(\cdot)$ and the backbone $f(\cdot)$, the goal of class-incremental learning is to train the model $\Phi(f(\cdot))$ on tasks $\{\mathcal{T}_{t}\}_{t=1}^n$ sequentially so that the model can classify testing samples from any task.
Specifically, at the $t$-th task $\mathcal{T}_t$, the training set is $\mathcal{D}_t = \{x^t_m, y^t_m\}_{m=1}^{N_t}$, where $x^t_m$ is the $m$-th image at task $\mathcal{T}_t$ with label $y_m^t$.
And the label space $\mathcal{Y}_t$ of task $\mathcal{T}_t$ is disjoint with other tasks, \ie $\bigcap_{t=1}^{n} \mathcal{Y}_t = \emptyset$.
Once the model finishes learning on task $\mathcal{T}_t$, the corresponding training set $\mathcal{D}_t$ will be dropped and becomes inaccessible when learning from $\mathcal{T}_{t+1}$, which raises a risk of forgetting old tasks when learning new ones.

\begin{figure*}
    \centering
      \includegraphics[width=0.85\linewidth]{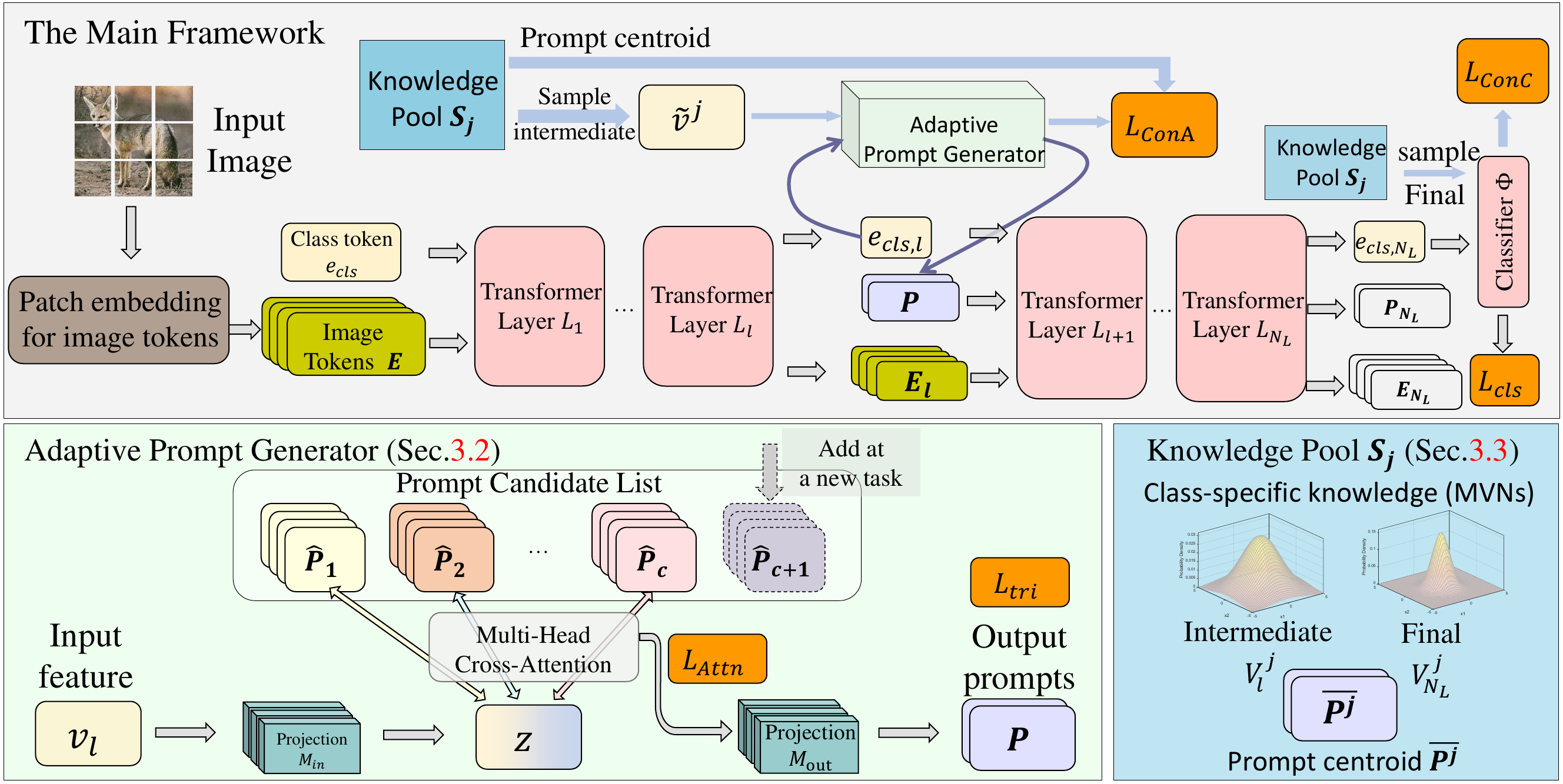}
      \caption{Overview of the proposed method. The Main framework is shown at the top part of the figure. The adaptive-prompting conducts in the middle of Layer $L_l$ and $L_{l+1}$. 
      With the help of the adaptive prompt generator (APG) which adaptively aggregates old and new knowledge, deeper layers are instructed by the
       old knowledge. 
      A corresponding illustration of the APG is shown in the lower-left corner of the figure. To facilitate the knowledge aggregation, the knowledge pool (shown in the lower-right corner) consists of class-specific knowledge and the prompt centroid is used to constrain the APG and the classifier $\Phi$ of the old knowledge. }
    \label{fig:method}
  \end{figure*}

\vspace{3pt}  
\noindent\textbf{Vision transformers and prompting.} \ \ 
Different from convolutional neural networks, Vision Transformers (ViTs) treat the image as a sequence of tokens.
Specifically, the image $x_m^t$ will be converted into a sequence $\mathbf{E} \in \mathbb{R}^{N_E\times d}$ through patch embedding layers, where $N_E$ is the total number of image patches (\ie tokens) and $d$ is the embedding dimension.
An extra token called \textit{class token} $e_{cls} \in \mathbb{R}^{d}$ will be concatenated to $\mathbf{E}$ to gather information from patches. 
The resultant token sequence $\mathbf{X} = \left[e_{cls};\mathbf{E}\right]$ will be fed into transformer layers for feature extraction, and the core is self-attention operation: 
    \begin{gather}
    \small
        \text{SelfAttn}(\mathbf{X}) = \text{Attn}(\mathbf{X}\mathbf{W}^Q , \mathbf{X}\mathbf{W}^K , \mathbf{X}\mathbf{W}^V ), \notag\\
        \text{Attn}(Q, K, V) = \text{Softmax}(\frac{QK^T}{\sqrt{d}})V,
    \end{gather}
where $\mathbf{W}^Q$, $\mathbf{W}^K$ and $\mathbf{W}^V$ are learnable projections.

To facilitate visual feature extraction, a common prompting~\cite{liu2021pre, liu2021gpt, P_tuning_v2, li2021prefix, vpt, liu2022prompt} technique is to insert a set of learnable extra tokens $\mathbf{P} \in \mathbb{R}^{N_P \times d}$ into the original sequence as $\left[e_{cls}; \mathbf{P} ;\mathbf{E}\right]$.
In this work, we only discuss vanilla ViTs~\cite{vit, deit} due to their simplicity and versatility.
We detail our method as follows, and an overview is in \cref{fig:method}.
\subsection{Adaptive Prompt Generation}
\label{sec:APG}
\noindent\textbf{Adaptive prompt generator.}\ \
We propose an adaptive prompt generator (APG) for adaptively aggregating old and new knowledge.
Assume there are $N_L$ transformer layers denoted as $\left\{ L_l \right\}_{l=1}^{N_L}$, and the output of the $l$-th transformer layer is $\mathbf{X}_l = L_l(L_{l-1}(\cdots L_1(x))) \in \mathbb{R}^{(N_E+1) \times d}$, where $x$\footnote{In the later part of \cref{sec:APG}, we will omit the task label $t$ and the index $m$ of image $x_m^t$ for simplicity.} is the input image.
The APG takes the intermediate feature as input, which is denoted as $v_l = \mathbf{X}_l[0, :]$, and outputs the class-specific prompts to facilitate feature extraction in deeper layers:
\begin{equation}
\mathbf{P} = \text{APG}(v_l) \in \mathbb{R}^{N_P \times d}.
\end{equation}

Specifically, in the APG, we keep updating a prompt candidate list to maintain different types of knowledge. 
Starting from an empty list denoted as $\mathbf{I}_0 = []$, the list is extended at each task, and the list at task $\mathcal{T}_t$ is denoted as:
\begin{equation}
    \small
    \mathbf{I}_{t} = \left[\mathbf{I}_{t-1}; \widehat{\mathbf{P}} _1, \ldots, \widehat{\mathbf{P}} _c, \ldots, \widehat{\mathbf{P}} _{\vert\mathcal{Y}_t\vert}\right],
\end{equation}
where $\widehat{\mathbf{P}} _c \in \mathbb{R}^{N_{g} \times d}$ represents a group of prompts candidates, and there are totally $N_{g} \times \vert \mathcal{Y}_t \vert$ prompt candidates are extended at $\mathcal{T}_t$ to form $\mathbf{I}_{t}$.
In order to generate prompts with class-specific knowledge adaptively for $v_l$, we firstly apply a projection module $M_{in}$ to $v_l$, denoted as $z = M_{in}(v_l)$, and then adopt a cross-attention operation between the projection $z$ and the candidate list $\mathbf{I}_{t}$:
\begin{equation}
    \widetilde{\mathbf{P}} = \text{CrossAttn}(z, \mathbf{I}_{t}) = \text{MMHA}(z, \mathbf{I}_{t}, \mathbf{I}_{t}),
\end{equation}
where the $\text{MMHA}$ denotes Multi-output Multi-head Attention, an extension of the Multi-head attention (MHA)~\cite{attention_is_all_you_need}.
While the original MHA outputs the same number of tokens with the input query, in our case, the query is a single token $z$ and we want to obtain $N_{P}$ prompts after cross-attention.
To avoid restricting $N_{P}=1$, we directly extend the MHA and form the MMHA as follows.
Firstly, with the input $z$ and $\mathbf{I}_t$, we define the $j$-th token computed from the $h$-th attention head as: 
\begin{equation}
\label{eq:cross_attn}
    r_h^{j} = \text{Attn}(z \mathbf{W}_{h,j}^Q, \mathbf{I}_{t} \mathbf{W}_{h,j}^K, \mathbf{I}_{t} \mathbf{W}_{h,j}^V).
\end{equation}
Assuming the number of heads is $n_h$, we concatenate results from every head as the $j$-th token $\textbf{R}_j$, and pass the token through a projection module as the output of MMHA:
\begin{equation}
\begin{aligned}
    \textbf{R}_j = \text{Concat}(r_1^j, \ldots, r_h^j, \ldots, r_{n_h}^j), \\
    \widetilde{\mathbf{P}} = \text{MMHA}(z, \mathbf{I}_{t}, \mathbf{I}_{t})  = \left[\mathbf{R}_1, \ldots, \mathbf{R}_{N_P}  \right]\mathbf{W}^o.
\end{aligned}
\end{equation}
We further put the $\widetilde{\mathbf{P}}$ into an output projection module $M_{out}$, and the obtained tokens $\mathbf{P} = M_{out}(\widetilde{\mathbf{P}}) \in \mathbb{R}^{N_P \times d}$ are the adaptive prompts for current image $x$.

The generated prompts $\mathbf{P}$ are further integrated with the original output $\mathbf{X}_l$ of the $l$-th layer, and the token sequence for the next layer becomes $\mathbf{X}_l = \left[e_{cls,l}; \mathbf{P} ;\mathbf{E}_l\right]$.
we feed $\mathbf{X}_l$ into the remaining deeper layers $\{L_{i}\}_{i=l+1}^{N_L}$.
The final output of the network with APG is denoted as $\mathbf{X}_{N_L}= \left[e_{cls,N_L}; \mathbf{P}_{N_L}; \mathbf{E}_{N_L}\right]$.
We take the class token $e_{cls,N_L}$ as the image feature.
The classifier $\Phi(\cdot)$ is composed of a fully-connected layer and a softmax function and outputs the classification probability $\Phi(e_{cls, N_L})$.

It is worth mentioning that all components inside the APG are trainable, including the projection module $M_{in},M_{out}$, the prompt candidate list $\mathbf{I}_{t}$ and the parameters in cross-attention operation ($\mathbf{W}^o$, $\{\mathbf{W}_{h,j}^{Q/K/V}\}$).

\vspace{3pt}
\noindent\textbf{Optimization of the APG.}\ \ 
The generated prompts are supposed to contain sufficient knowledge so that the model can handle the current task. 
Hence, we adopt a classification loss to learn new knowledge from the current task:
\begin{equation}
    \mathcal{L}_{cls} = - \log\Phi(e_{cls, N_L})^{(y)},
\end{equation}
where $\Phi(e_{cls, N_L})^{(y)}$ is the $y$-th element of $\Phi(e_{cls, N_L})$.

Since $\mathcal{L}_{cls}$ is evaluated on the ultimate feature representation $e_{cls, N_L}$, we further directly apply constraints on the generated prompts and the prompt candidate list for more effective learning.  
For the prompt candidates, its information is aggregated through the cross-attention operation in Eq. \ref{eq:cross_attn}.
When computing the $j$-th token $r_h^{j}$ in the $h$-th attention head, the attention score (Eq. \ref{eq:cross_attn}) is calculated as:
\begin{equation}
    A_h^j = \text{Softmax}(\frac{(z\mathbf{W}_{h,j}^Q)(\mathbf{I}_{t}\mathbf{W}_{h,j}^K)^T}{\sqrt{d}}),
\end{equation}
where each element $A_h^{j,(c)}$ indicates the score between the projection $z$ and the $c$-th prompt in the candidate list $\mathbf{I}_{t}$ under the projection $\mathbf{W}_{h,j}^Q, \mathbf{W}_{h,j}^K$.
The prompt candidates should contain diverse knowledge from each class, otherwise, the APG can not generate prompts adaptively to various images.
In other words, if the candidates only learn the knowledge from part of the classes, APG can not handle images from the other classes.
Thus, we explicitly guide distinctive groups of prompt candidates to learn class-specific knowledge. We apply a class-specific constraint on the attention score:
\begin{equation}
    L_{attn} = -\sum_{j=1}^{n_p}\sum_{h=1}^{n_h} \sum_{c \in \widehat{C}_y} \log(A_h^{j,(c)}),
\end{equation}
where $\widehat{C}_y$ is the set of indexes of the prompts in $\widehat{P}_{y}$.

Moreover, since the images of the same class have similar characteristics, such as appearance, it is natural that the network can be prompted in a similar way when extracting features for different images in a class.
To this end, we further constrain the relations between prompts through a triplet loss, encouraging the APG to learn the common knowledge of each class.
Assume $(x_{1},x_{2},x_{3})$ is a triplet of images and their labels satisfy $y_{1} = y_{2}, y_{1} \neq y_{3}$.
The corresponding prompts generated by the APG are $\mathbf{P}_{1}$, $\mathbf{P}_{2}$ and $\mathbf{P}_{3}$.
Since $x_1$ and $x_2$ are from the same class while $x_3$ is from another class, we constrain the distance between $\mathbf{P}_{1}$ and $\mathbf{P}_{2}$ should be smaller than that between  $\mathbf{P}_{1}$ and $\mathbf{P}_{3}$.
Formally, the cosine distance between $\mathbf{P}_{1}$ and $\mathbf{P}_{2}$ is denoted as $d_p = \cos(\mathbf{P}_{1}, \mathbf{P}_{2})$, and similarly we have $d_n = \cos(\mathbf{P}_{1}, \mathbf{P}_{3})$. The triplet loss with margin is adopted:
\begin{equation}
    \mathcal{L}_{tri} = [d_p - d_n + \alpha]_+,
\end{equation}
where $[\cdot]_+$ denotes the hinge loss, and $\alpha$ is the margin.

\vspace{3pt}
\noindent\textbf{Discussion.}\ \ 
The proposed APG has the following advantages. 
First, APG continuously learns to generate diverse prompts conditioned on the input of the network.
Hence, APG breaks the limitation of selecting prompts from a fixed-size prompt pool and alleviates the dependence on the pretrained model that is used for querying discrete prompts from a pool.
Second, the prompt candidate list is extendable to handle growing knowledge. This enables the model capable of long incremental learning (see \cref{sec:non_pretrianed} and \cref{sec:pretrained}) as the APG can aggregate knowledge from different tasks by the cross-attention operation (Eq. \ref{eq:cross_attn}).

\subsection{Anti-forgetting Learning: A Knowledge Pool}

\noindent\textbf{Construction of the knowledge pool.}\ \ 
Any trainable module faces the forgetting (\ie old knowledge degradation) problem when learning from sequential tasks, including the proposed APG and the classifier $\Phi(\cdot)$.
Hence, we develop a knowledge pool to better maintain old knowledge.

Once finishing learning at task $\mathcal{T}_t$, we summarize the knowledge in the training set $\mathcal{D}_t$ into the knowledge pool before dropping the dataset.
Specifically, denote all the images in class $c$ as $\mathcal{D}_t^c = \{x_m^{t}, y_m^{t} \vert x_m^{t} \in \mathcal{D}_t, y_m^{t}=c  \}$.
We feed these images into the backbone and get a set of features $V_c^l = \{v_m^{l} \vert x_m^{t} \in \mathcal{D}_t^c \}$ from the $l$-th transformer layer.
Then the class centroid $\mu^l_{c}$  is then calculated by averaging features and the and $\Sigma^l_c$ is a matrix where each element $\Sigma_c^{l,(i,k)}$ is the covariance between the $i$-th feature and the $k$-th feature. We use the class-specific statistics to form a multivariate normal distribution $\mathcal{N}^l_c=\mathcal{N}(\mu^l_c, \Sigma^l_c)$ for each class.
Besides, to better constrain the APG, the feature centroid $\mu^l_c$ is used to extract corresponding prompts by feeding the centroid into the APG, \ie $\overline{\mathbf{P}^{c}} = \text{APG}(\mu_c)$.

Finally, we form the knowledge pool for class $c$ as $\mathcal{S}_c = (\mathcal{N}^l_c, \mathcal{N}^{N_L}_c, \overline{\mathbf{P}^{c}})$, where $\mathcal{N}^{N_L}_c$ is computed from the features from the last layer.
Generally, we store the statistics for each class $\{\mathcal{S}_c\}_{c=1}^{N_t^{cls}}$, where $N_t^{cls} = \sum_t \vert \mathcal{Y}_t \vert $ is the number of seen classes so far.

\vspace{3pt}
\noindent\textbf{Anti-forgetting learning.}\ \ 
For exemplar-free incremental learning, we can only access the current training set $\mathcal{D}_t$ during task $\mathcal{T}_t$.
The knowledge pool $\{\mathcal{S}_c\}_{c=1}^{N_{t-1}^{cls}}$ described above is used to maintain old knowledge.
To alleviate forgetting in the APG, we sample from $\mathcal{N}^l_c$, and get intermediate knowledge vector $\widetilde{v}^c \sim \mathcal{N}^l_c$.
With the vector and the corresponding prompt centroid $\overline{\mathbf{P}^{j}}$, we apply a constraint on APG:
\begin{equation}
    \mathcal{L}_{ConA} = \phi(\text{APG}(\widetilde{v}^c), \overline{\mathbf{P}^{c}}),
\end{equation}
where $\phi$ is a distance metric and we find that the $L1$ loss is useful.
With the constraint, APG can be reminded of old knowledge and perform aggregation on them.

The classifier $\Phi(\cdot)$ is extended after each task, and needs to be updated during every task.
Thus forgetting also happens at the classifier.
Similar to the constraint on the APG, we sample the knowledge vector from $\widetilde{v}^c \sim \mathcal{N}^{N_L}_c$.
Then a simple but effective cross-entropy loss is adopted:
\begin{equation}
    \mathcal{L}_{ConC} = -\log\Phi(\widetilde{v}_m^c)^{(c)},
\end{equation}
where $\Phi(\widetilde{v}^c)^{(c)}$ is the $c$-th element of $\Phi(\widetilde{v}^c)$.

\vspace{3pt}
\noindent\textbf{Training objective for the whole model.}
The objective function for optimizing the model and APG is summarized as follows:
\begin{equation}
    \mathcal{L} = \mathcal{L}_{attn} + \mathcal{L}_{tri} + \mathcal{L}_{cls} + \mathcal{L}_{conA} + \mathcal{L}_{conC}.
\end{equation}
For the backbone $f(\cdot)$, it is only trained on the first task and then is fixed in the case of without pretraining, and otherwise it is fixed all the time. The classifier $\Phi(\cdot)$ and APG are optimized across all tasks.

\begin{table*}[t]
\footnotesize
\centering
\begin{tabular}{c|cc|cc|cc|cc}
\bottomrule[1.2pt]
\multirow{2}{*}{Methods} & \multicolumn{2}{c|}{\begin{tabular}[c]{@{}c@{}}ImageNet-Sub\\ B50-T10\end{tabular}} & \multicolumn{2}{c|}{\begin{tabular}[c]{@{}c@{}}ImageNet-Sub\\ B30-T14\end{tabular}} & \multicolumn{2}{c|}{\begin{tabular}[c]{@{}c@{}}CIFAR100\\ B50-T10\end{tabular}} & \multicolumn{2}{c}{\begin{tabular}[c]{@{}c@{}}CIFAR100\\ B40-T20\end{tabular}} \\ \cline{2-9} 
 & Avg. Acc.↑ & Forgetting↓ & Avg. Acc.↑ & Forgetting↓ & Avg. Acc.↑ & Forgetting↓ & Avg. Acc.↑ & Forgetting↓ \\ \hline
L2P & 25.11 & 1.72 & 29.92 & 3.03 & 36.55 & 2.61 & 18.84 & 3.02 \\
DualPrompt & 35.82 & 4.53 & 30.33 & 4.16 & 26.84 & 4.68 & 11.86 & 1.80 \\ \hline
Ours & \textbf{75.52} & 5.63 & \textbf{69.87} & 6.87 & \textbf{66.68} & 5.42 & \textbf{62.41} & 6.72 \\
\toprule[1.2pt]
\end{tabular}
\caption{Comparison with state-of-the-art prompt-based methods.
    Since existing prompt-based methods rely on a pretrained backbone, we treat the first task (\ie 50 classes for the B50 setting)
    as the pretraining task and we conduct class-incremental learning on the rest classes using their methods.
    The best results are marked in \textbf{bold}.
    Results are averaged across three trials.
    }
\label{table:l2p_dualprompt_non_pretrained}
\end{table*}

\section{Experiments}
\label{sec:Exp}

\begin{table}[t]
\footnotesize
\centering
\begin{tabular}{c|c|ccc}

\bottomrule[1.2pt]
       &       & B50-T10 & B30-T14 & B50-T50 \\\cline{3-5} 
\multirow{-2}{*}{Methods} & \multirow{-2}{*}{\begin{tabular}[c]{@{}c@{}}Mem \\ Per cls\end{tabular}} & \multicolumn{3}{c}{Avg.   Acc.(\%)}      \\\hline
Imagine~\cite{Imagine} & 20          & 76.76      & -          & -          \\
PODNet~\cite{podnet}   & 20    & 74.58      & 68.83      & 62.48      \\
UCIR~\cite{ucir}       & 20    & 67.82      & -          & 57.25          \\
DDE~\cite{DDE}         & 20    & 75.41      & 65.11      & -      \\
CwD~\cite{cwd}         & 20    & 76.91      & -          & -          \\
AFC~\cite{afc}         & 20    & 75.75      & 72.47      & 71.97      \\
Foster~\cite{foster}   & 20    & 75.72      & 75.28      & 69.76       \\ \hline
Imagine~\cite{Imagine} & 10    & 74.94      & -          & -          \\
PODNet~\cite{podnet}   & 10    & 70.40      & 60.18      & 41.47      \\
UCIR~\cite{ucir}       & 10    & 64.04      & -          & -          \\
DDE~\cite{DDE}         & 10    & 73.00      & -          & -          \\
AFC~\cite{afc}         & 10    & 74.85      & 70.46      & 63.28      \\
Foster~\cite{foster}   & 10    & 71.46      & 65.91      & 58.79      \\ \hline
EWC~\cite{ewc}         & 0     & 20.40      & 18.43      & 5.9      \\
LwF\_MC~\cite{lwf}     & 0     & 31.18      & 13.22      & -          \\
MUC~\cite{muc}         & 0     & 35.07      & -          & -          \\
SDC~\cite{SDC}         & 0     & 61.12      & -          & -          \\
PASS~\cite{PASS}       & 0     & 61.80      & -          & -          \\
SSRE~\cite{SSRE}       & 0     & 67.69      & -          & -          \\
\rowcolor[HTML]{D9D9D9} 
Ours    & 0   & \textbf{75.52}      & \textbf{69.87}      & \textbf{75.70} \\
\toprule[1.2pt]
\end{tabular}
\caption{Comparison with state-of-the-art methods trained from scratch on ImageNet-SubSet. 
We list rehearsal-based methods with 20 images per class and their variant with 10 images.
The best results among exemplar-free methods are marked in \textbf{bold}.
Results are averaged across three trials.
}
\label{table:imagenet}
\end{table}

\subsection{Experiment Setup}
\label{sec:exp_setup}
\noindent\textbf{Datasets.} 
We conduct experiments on three datasets: CIFAR100~\cite{cifar100}, ImageNet-Subset~\cite{imagenet}, and ImageNet-R~\cite{imagenerR}.
CIFAR100~\cite{cifar100} contains 100 classes.
There are 50,000 images for training and 10,000 images for evaluation. 
ImageNet~\cite{imagenet} is a large-scale dataset containing 1,000 classes, 1300 images per class. 
We follow prior works~\cite{ucir,podnet, DDE, der, bic, dmc, inthewild, Imagine, foster} and use ImageNet-Subset which contains 100 classes.
ImageNet-R~\cite{imagenerR} dataset contains newly collected data that are different styles of the original ImageNet classes.
We follow prior work~\cite{dualprompt} to split the dataset with 80\% samples for training and the rest 20\% samples are used for evaluation.

\vspace{3pt}
\noindent\textbf{Training and testing protocols.} 
For the training and testing protocols, we follow previous works~\cite{podnet, ucir, der, DDE, Imagine, foster, l2p,dualprompt} to conduct class-incremental learning.
After every training stage, we evaluate the model by testing on the union of all testing sets of seen tasks. 
The classification accuracies in every task are averaged, and we report the \textit{average accuracy} in all experiments~\cite{icarl,podnet,der,DDE,bic,dualprompt,l2p,Imagine}.
When comparing with L2P~\cite{l2p} and DualPrompt~\cite{dualprompt}, we also report the \textit{forgetting} metric following their papers.

We do not save any images as a memory buffer and only keep the class statics of old classes on the feature level, following the exemplar-free class-incremental learning proposed in PASS~\cite{PASS} and SSRE~\cite{SSRE}. 

\vspace{3pt}
\noindent\textbf{Implementation details.} 
For all experiments below, we use the vanilla ViT as our backbone. 
For the standard non-pretrained setting, we adjust the embedding dimension and number of heads to match the number of parameters of the widely used ResNet-18~\cite{resnet}. For more information about the adjusted backbone, please refer to the appendix.
We follow the training strategy proposed in DeiT(without distillation)~\cite{deit}, using AdamW as the optimizer with initial learning of 5e-4.
For class-incremental learning with a pretrained backbone, we follow prior works\cite{l2p, dualprompt} using ViT-Base as our backbone.

\vspace{3pt}
\noindent\textbf{Setting notation.} We denote class-incremental learning settings in form of: `B$x$-T$y$', which means the first task contains $x$ classes, and the rest classes are evenly divided into $y$ tasks to incrementally learn. For example, `B50-T10' means the first task contains 50 classes and the rest classes are divided into 10 tasks evenly.

\subsection{Non-pretrained Class-incremental Learning}
\label{sec:non_pretrianed}
In this section, we evaluate our method under the standard non-pretrained incremental learning setting\cite{der, podnet, ucir, DDE, cwd, afc,foster,dytox,ewc,lwf,muc,SDC,PASS,SSRE}.

\vspace{3pt}
\noindent\textbf{Comparison with prompted-based methods.} 
In the popular non-pretrained scenario, we first compare our method with prompted-based methods L2P~\cite{l2p} and DualPrompt~\cite{dualprompt}.
Since existing prompt-based methods rely on a pretrained backbone to conduct incremental learning, for the non-pretrained setting, we treat the first task during incremental learning as a `pretraining' task and train a backbone in a supervised manner.
Later, we re-implement L2P and DualPrompt according to the official code and conduct incremental learning on the rest tasks based on the pretrained weight.
In this way, we can study the situation when future tasks are totally disjoined from the pre-trained task.

The experiment results are shown in \Cref{table:l2p_dualprompt_non_pretrained}. 
In terms of average accuracy, our proposed method outperforms L2P and DualPrompt by a large margin under different settings of the number of tasks.
The forgetting metric of L2P and DualPrompt is incredibly low.
We found they basically maintain the first task's performance and perform poorly on new tasks.
Therefore the forgetting metric (which can be regarded as an average \textit{maximum accuracy drop}) is low. 
Nevertheless, our method still achieves competitive results on forgetting compared to these two methods.
We validate this observation by varying the number of classes at the first task and the total number of tasks on two different datasets.

\begin{table}[t]
    \footnotesize
    \centering
    \tabcolsep=0.15cm
\begin{tabular}{c|c|cccc}
    \bottomrule[1.2pt]
            &           & B50-T10 & B50-T5 & B40-T20 & B30-T14 \\\cline{3-6} 
    \multirow{-2}{*}{Methods} & \multirow{-2}{*}{\begin{tabular}[c]{@{}c@{}}Mem \\ Per cls\end{tabular}} & \multicolumn{4}{c}{Avg.   Acc.(\%)} \\\hline
    Imagine\cite{imagenet}        &20         & 66.47      & 68.01     &  -         &  - \\
    PODNet~\cite{podnet}           & 20        & 66.70      & 62.65     & 58.92      & 53.91           \\
    UCIR~\cite{ucir}             & 20        & 62.77      & 56.75     & 42.88      & 45.07       \\
    CwD~\cite{cwd}              & 20        & 70.30      & -         & -          & -           \\
    AFC~\cite{afc}              &20         &64.98       &66.49           &62.19            &72.47            \\
    DDE~\cite{DDE}              &20         &65.42       & 64.12          & -           &   -         \\
    Foster~\cite{foster}           & 20        & 77.54      & 67.69     & 56.87      & 53.80       \\\hline
    Imagine\cite{Imagine}          & 10        & 64.41      & 67.08     & -          & -           \\         
    PODNet~\cite{podnet}           & 10        & 60.09      & 62.03     & 55.78      & 55.93       \\
    UCIR~\cite{ucir}             & 10        & 56.95      & 59.95     & 42.88      & 45.07       \\
    AFC~\cite{afc}              & 10        & 61.77      & 63.98     & 57.06      & 58.81         \\
    DDE~\cite{DDE}              &10         & 64.41      & 61.47          &-            &  -          \\
    Foster~\cite{foster}           & 10        & 57.70      & 60.51     & 33.18      & 54.57       \\\hline
    EWC~\cite{ewc}              & 0         & 24.48      & 21.20     & 15.89      & 15.89       \\
    LwF\_MC~\cite{lwf}          & 0         & 45.93      & 27.43     & 20.07      & 21.05       \\
    MUC~\cite{muc}              & 0         & 49.42      & 30.19     & 21.27      & -           \\
    SDC~\cite{SDC}              & 0         & 56.77      & 57.00     & 58.90      & -           \\
    PASS~\cite{PASS}             & 0         & 63.47      & 61.84     & 58.09      & 56.61       \\
    SSRE~\cite{SSRE}             & 0         & 65.88      & 65.04     & 61.70      & -           \\
    \rowcolor[HTML]{D9D9D9} 
    Ours             & 0             & \textbf{66.68}      & \textbf{66.83}      & \textbf{62.41}      & \textbf{57.57}  \\    
    \toprule[1.2pt]
    \end{tabular}
    \caption{Comparison with state-of-the-art methods trained from scratch on CIFAR100. 
    We list rehearsal-based methods with 20 images per class and their variant with 10 images.
    The best results are marked in \textbf{bold}.
    Results are averaged across three trials.
    }
    \label{table:CIFAR100}
    \end{table}    

\vspace{3pt}
\noindent\textbf{Comparison with exemplar-free methods.}
To compare the proposed method with pioneering exemplar-free\cite{SDC, PASS, SSRE, lwf} works fairly, we conduct experiments on ImageNet-Subset and CIFAR100 under different numbers of tasks.
The detailed results on ImageNet-Subset are listed in \Cref{table:imagenet}.
For the 10-task setting, we find that our method outperforms other exemplar-free methods by a large margin.
Specifically, our method achieves 75.52\% average accuracy, which is 7.95\%, 13.84\%, 14.52\% higher than the state-of-the-art methods SSRE~\cite{SSRE}, PASS~\cite{PASS} and SDC~\cite{SDC}.
On CIFAR100, our method still performs better than other exemplar-free methods(see \Cref{table:CIFAR100}).
Specifically, our method outperforms the exemplar-free method SSRE by 2.56\%, 1.65\%, and 0.66\% on average accuracy under 5, 10 and 20 task settings.
\vspace{3pt}
\noindent\textbf{Comparison with rehearsal-based methods.}
To show the effectiveness of our adaptive-prompting scheme, we further report the results of rehearsal-based methods on ImageNet-Subset and CIFAR100 under different numbers of tasks and different sizes of memory budgets. 

The comparison results of different settings on ImageNet-Subset are shown in \Cref{table:imagenet}. 
Our method (w/o exemplars) achieves 75.52\% Avg. Acc. at B50-T10 setting, outperforming PODNet~\cite{podnet} and AFC~\cite{afc} (both 20 exemplars per class) by 1.06\% and 0.23\%.
To explore the effectiveness of our proposed method under a long incremental process, we also conduct experiments of B50-T50 (\ie 50-task incremental learning).
When the learning process gets longer from 10 tasks to 50 tasks, our performance maintains around 75\% and outperforms other methods by a large margin.
This is mainly because our proposed extendable candidate list in the APG enables our method to learn in long incremental tasks while other methods suffer greatly from the forgetting problem.
When we tighten the memory budget, rehearsal-based methods face a drastic performance drop.
Compared with Foster~\cite{foster} and AFC~\cite{afc} (both kept 10 images per class) under the B50-T10 setting, our method achieves 4.56\% and 0.67\% higher average accuracy. 
Even for a method like Imagine~\cite{Imagine} which requires an extra auxiliary dataset to diversify the limited memory,
our method still outperforms it by 0.7\% accuracy when it kept 10 images per class.
More results on CIFAR100 are in Table \ref{table:CIFAR100}.

\begin{table}[t]
\centering
\footnotesize
\tabcolsep=0.18cm
\begin{tabular}{ccccccc}
\bottomrule[1.2pt]
\multicolumn{1}{c|}{\multirow{2}{*}{Methods}} & \multicolumn{2}{c|}{\begin{tabular}[c]{@{}c@{}}CIFAR100\\ B0-5\end{tabular}} & \multicolumn{2}{c|}{\begin{tabular}[c]{@{}c@{}}CIFAR100\\ B0-10\end{tabular}} & \multicolumn{2}{c}{\begin{tabular}[c]{@{}c@{}}CIFAR100\\ B0-20\end{tabular}} \\ \cline{2-7}
\multicolumn{1}{c|}{} & A$\uparrow$  & \multicolumn{1}{c|}{F$\downarrow$} & A$\uparrow$  & \multicolumn{1}{c|}{F$\downarrow$} & A$\uparrow$  & F$\downarrow$ \\ \hline
\multicolumn{7}{c}{\textit{ImageNet-21K Pretraining}} \\ \hline
\multicolumn{1}{c|}{Upper Bound} & 90.80 & \multicolumn{1}{c|}{-} & 91.22 & \multicolumn{1}{c|}{-} & 91.53 & - \\
\multicolumn{1}{c|}{FT} & 43.30 & \multicolumn{1}{c|}{51.02} & 28.38 & \multicolumn{1}{c|}{69.31} & 17.63 & 84.32 \\ \hline
\multicolumn{1}{c|}{L2P\cite{l2p}} & 88.65 & \multicolumn{1}{c|}{88.64} & 88.64 & \multicolumn{1}{c|}{7.35} & 86.57 & 9.31 \\
\multicolumn{1}{c|}{DualPrompt\cite{dualprompt}} & \textbf{89.56} & \multicolumn{1}{c|}{\textbf{89.12}} & 89.12 & \multicolumn{1}{c|}{\textbf{5.16}} & 87.61 & 6.92 \\
\rowcolor[HTML]{D9D9D9} 
\multicolumn{1}{c|}{Ours} & 88.48 & \multicolumn{1}{c|}{89.35} & \textbf{89.35} & \multicolumn{1}{c|}{6.01} & \textbf{88.64} & \textbf{6.51} \\ \hline
\multicolumn{7}{c}{\textit{Tiny-ImageNet Pretraining}} \\ \hline
\multicolumn{1}{c|}{Upper Bound} & 89.42 & \multicolumn{1}{c|}{-} & 89.37 & \multicolumn{1}{c|}{-} & 88.44 & - \\
\multicolumn{1}{c|}{FT} & 43.16 & \multicolumn{1}{c|}{50.95} & 28.36 & \multicolumn{1}{c|}{69.61} & 17.65 & 84.25 \\ \hline
\multicolumn{1}{c|}{L2P\cite{l2p}} & 84.68 & \multicolumn{1}{c|}{9.49} & 82.51 & \multicolumn{1}{c|}{7.78} & 78.69 & 8.12 \\
\multicolumn{1}{c|}{DualPrompt\cite{dualprompt}} & 84.54 & \multicolumn{1}{c|}{8.25} & 81.71 & \multicolumn{1}{c|}{9.70} & 79.13 & 11.18 \\
\rowcolor[HTML]{D9D9D9} 
\multicolumn{1}{c|}{Ours} & \textbf{87.72} & \multicolumn{1}{c|}{\textbf{7.56}} & \textbf{87.61} & \multicolumn{1}{c|}{\textbf{7.16}} & \textbf{87.50} & \textbf{7.49}\\
\toprule[1.2pt]
\end{tabular}
\caption{
Comparison to state-of-the-art prompt-based methods with ImageNet-21k/Tiny-IamgeNet pretrained weights on CIFAR100.
The best and the second best results are marked in \textbf{bold}.
`A' means Avg. Acc and `F' indicates the forgetting metric.
Our method effectively alleviates reliance on the strong pretraining (Tiny-ImageNet pretraining) and can still benefit from pretraining (ImageNet-21k pretraining).
}
\label{table:CIFAR100_different_pretrain}
\end{table}

\subsection{Evaluation with Pretrained Backbone} 
\label{sec:pretrained}
\noindent\textbf{Incremental learning with intensive pretrained backbone.}
To demonstrate the flexibility of the proposed method, we further extend our method to the setting with pretrained weight and conduct experiments with the pretrained weight as previous methods~\cite{l2p, dualprompt} did.
We follow previous works~\cite{l2p,dualprompt} to load an ImageNet 21k pretrained weight~\cite{deit} for the vanilla ViT-Base/16. 
We conduct experiments on two different datasets CIFAR100 and ImageNet-R following DualPrompt~\cite{dualprompt}.
For each dataset, we conduct 5, 10, and 20 incremental tasks to validate different methods.
Furthermore, we also conduct an upper bound and a finetuning baseline (FT) for reference.
An upper bound is constructed by treating all training samples available across tasks and turning the pretrained ViT backbone using supervised training strategies.
The FT baseline is to turn the whole backbone (also initialized with the pretrained weight) without any constraint on the old knowledge during the learning.

As shown in \Cref{table:CIFAR100_different_pretrain}, our method makes comparable performance when the number of tasks is 5.
When the learning sequence becomes longer, our method outperforms L2P and DualPrompt by a larger margin.
Specifically, our proposed method outperforms DualPrompt and L2P by 1.03\% and 2.07\% on average accuracy on 20-task incremental learning on CIFAR100.
A similar conclusion can be observed under a more challenging dataset ImageNet-R in \Cref{table:pretrain_imagenetr}. 
As the training procedure gets longer, the superiority of our model begins to emerge.
On 20-task incremental learning on ImageNet-R, we achieve 0.74\% and 3.06\% improvement compared with DualPrompt at average accuracy and forgetting respectively.

It is worth mentioning that our method does not rely on a complex and delicate prompting pattern (\ie focus on where or how to insert prompts), only utilizing the vanilla way~\cite{vpt} to insert the prompts.
Experiments in this section show that adaptive-prompting is an effective design and the static selection process from a prompt pool in previous methods~\cite{l2p, dualprompt} can be the limitation of prompt-based methods.

\begin{table}[t]
\centering
\footnotesize
\tabcolsep=0.18cm
\begin{tabular}{c|cc|cc|cc}
\bottomrule[1.2pt]
\multicolumn{1}{c|}{\multirow{2}{*}{Methods}} & \multicolumn{2}{c|}{\begin{tabular}[c]{@{}c@{}}ImageNet-R\\ B0-T5\end{tabular}} & \multicolumn{2}{c|}{\begin{tabular}[c]{@{}c@{}}ImageNet-R\\ B0-T10\end{tabular}} & \multicolumn{2}{c}{\begin{tabular}[c]{@{}c@{}}ImageNet-R\\ B0-T20\end{tabular}} \\ \cline{2-7}
\multicolumn{1}{c|}{} & A$\uparrow$  & \multicolumn{1}{c|}{F$\downarrow$} & A$\uparrow$  & \multicolumn{1}{c|}{F$\downarrow$} & A$\uparrow$  & F$\downarrow$ \\ \hline
\multicolumn{1}{c|}{Upper Bound} & 77.47 & \multicolumn{1}{c|}{-} & 77.87 & \multicolumn{1}{c|}{-} & 78.90 & - \\
\multicolumn{1}{c|}{FT} & 37.73 & \multicolumn{1}{c|}{46.23} & 24.76 & \multicolumn{1}{c|}{63.14} & 18.66 & 72.35 \\ \hline
\multicolumn{1}{c|}{L2P\cite{l2p}} & 66.86 & \multicolumn{1}{c|}{5.02} & 66.61 & \multicolumn{1}{c|}{9.37} & 64.08 & 8.64 \\
\multicolumn{1}{c|}{DualPrompt\cite{dualprompt}} & \textbf{73.02} & \multicolumn{1}{c|}{3.73} & 72.57 & \multicolumn{1}{c|}{4.68} & 70.48 & 7.47 \\
\rowcolor[HTML]{D9D9D9} 
\multicolumn{1}{c|}{Ours} & 72.36 & \multicolumn{1}{c|}{6.37} & \textbf{73.27} & \multicolumn{1}{c|}{8.59} & \textbf{71.22} & 7.39 \\
\toprule[1.2pt]
\end{tabular}
\caption{Comparison with state-of-the-art methods with ImageNet-21k pretrained weights on ImageNet-R. 
   The best results are marked in \textbf{bold}.
   `A' means Avg. Acc and `F' indicates the forgetting metric.
       Even with the use of strong pretraining (\Cref{table:imagenet}), our method also achieves comparable performance to other methods.
       }
    \label{table:pretrain_imagenetr}
\end{table}

\noindent\textbf{Further analysis on different pretraining weights.}
To investigate the effect of the gap between pretrained knowledge and future tasks, we further conduct experiments on CIFAR100 using TinyImageNet pretrained weights~\cite{tinyimagenet_vit}.
As shown in \Cref{table:CIFAR100_different_pretrain}, we can observe that our method outperforms L2P by 3.04\%, 5.10\% and 8.81\% on Avg. Acc. under 5, 10 and 20 tasks.
Across different pretraining strategies (ImageNet-21k and TinyImageNet in \Cref{table:CIFAR100_different_pretrain} and standard non-pretrained in \Cref{table:l2p_dualprompt_non_pretrained}), our method achieves 1.03\%, 8.37\% and 39.54\% higher Avg. Acc. compared with the DualPrompt under the 20-task setting.
It is mainly because the pretrained knowledge dominates the incremental performance for existing prompt-based methods, making their methods fail when the semantic gap between the pretrained task and incremental tasks is relatively large.
By contrast, our method breaks the reliance on the strong pretrained backbone since the APG can effectively learn to reduce the negative effect of the semantic gap.

 \begin{table}[t]
 \centering
 \footnotesize
\begin{tabular}{c|cc|ccc|c}
\bottomrule[1.2pt]
\multirow{2}{*}{Configs} & \multicolumn{2}{c|}{w/o APG} & \multicolumn{3}{c|}{w/ APG} & \multirow{2}{*}{\begin{tabular}[c]{@{}c@{}}Avg. \\ Acc.(\%)\end{tabular}} \\
 & $\mathcal{L}_{cls}$ & $\mathcal{L}_{conC}$ & $\mathcal{L}_{conA}$ & $\mathcal{L}_{attn}$ & $\mathcal{L}_{tri}$ &  \\ \hline
c-1 & $\surd$ &  &  &  &  & 19.37 \\
c-2 & $\surd$ & $\surd$ &  &  &  & 71.84 \\ \hline
c-3 & $\surd$ & $\surd$ & $\surd$ &  &  & 73.46 \\
c-4 & $\surd$ & $\surd$ & $\surd$ & $\surd$ &  & 74.70 \\
c-5 & $\surd$ & $\surd$ & $\surd$ &  & $\surd$ & 75.59 \\
c-6 & $\surd$ & $\surd$ &  & $\surd$ & $\surd$ & 73.60 \\
Full & $\surd$ & $\surd$ & $\surd$ & $\surd$ & $\surd$ & \textbf{75.94} \\ \toprule[1.2pt]
\end{tabular}
\caption{Effectiveness of each loss function of the proposed method. `c-x' denotes different training configs. `Full' indicates the default full model. Experiments are conducted on ImageNet-Sub under non-pretrained 10-task setting. The best result is marked in \textbf{bold}.
}
\label{table:ab}
\end{table}

\subsection{Ablation Studies}
\label{sec:exp_ab}
In this section, we discuss the effectiveness of each loss function proposed in \cref{sec:method}.
We perform ablation studies on the ImageNet-Subset with a vanilla non-pretrained ViT backbone under the 10-task incremental learning setting.

\vspace{3pt}
\noindent\textbf{The effectiveness of knowledge pool.} Our knowledge pool is served in two aspects: one is the constraint on the APG ($\mathcal{L}_{conA}$) and the second is to guide the classifier ($\mathcal{L}_{conC}$).
In \Cref{table:ab}, c-1 indicates we only apply the classification loss $\mathcal{L}_{cls}$ and the performance is far from satisfactory because the network is inaccessible to any old knowledge.
We can observe that with the constraint $\mathcal{L}_{conC}$ on the classifier, the unbalance problem of the classifier~\cite{bic} can be greatly alleviated (Avg. Acc. from 19.37\% to 71.84\%).
This indicates that the knowledge vector can represent the original classes well, which can effectively overcome forgetting.
With the help of $\mathcal{L}_{conA}$, the APG is able to output effective prompts for the old classes.
The average accuracy is boosted from 71.84\% to 73.46\%.
This is mainly because the trainable module APG is facing severe forgetting problems during learning which can be effectively addressed by $\mathcal{L}_{conA}$.

\vspace{3pt}
\noindent\textbf{The effectiveness of the APG and the associated losses.} 
To further alleviate the forgetting problem, we design two losses in \cref{sec:method} to guide the APG to learn class-specific knowledge while training.
The first one is $\mathcal{L}_{attn}$, which guides the attention operation in the APG to be more class-related.
With the $\mathcal{L}_{attn}$ loss, the performance increases from 73.46\% to 74.70\%.
The triplet $\mathcal{L}_{tri}$ serves as a constraint that avoids the APG to generate degraded prompts and narrows the distance between two prompts with the same classes, bringing a performance boost of 1.24\% (compares c-4 with the full model).
\begin{figure}
    \centering
    \subfigure[The impact of which layers that APG augments into.]{
    \includegraphics[width=0.45\linewidth]{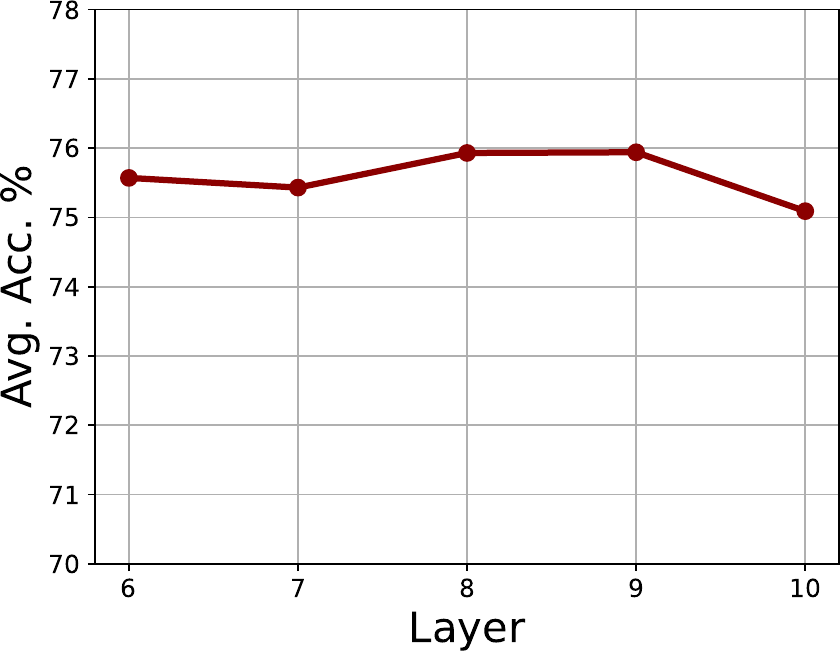}
    \label{fig:layer}
    }
    \hfill
    \subfigure[\fontsize{5bp}{13bp}The impact of the number of generated prompts.]{
    \includegraphics[width=0.45\linewidth]{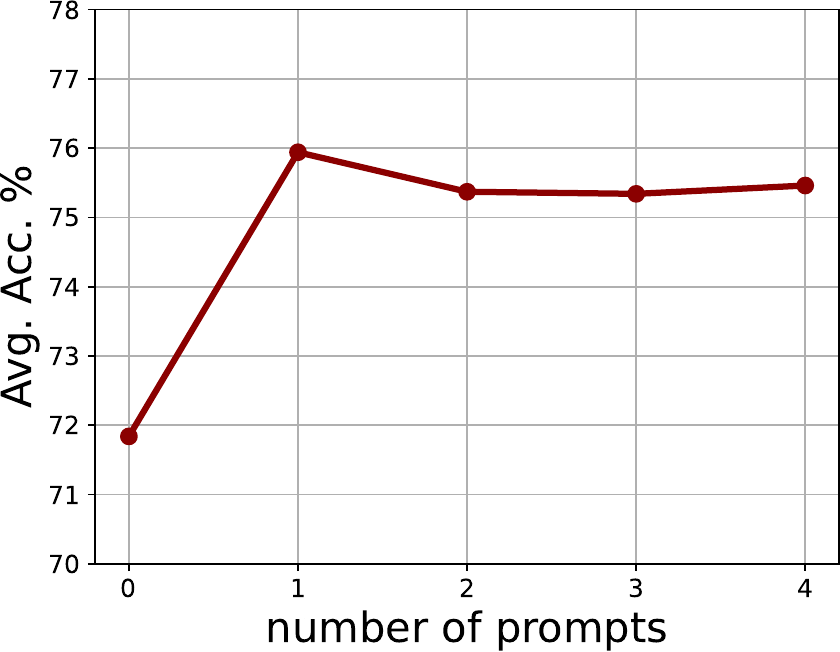}
    \label{fig:num_prompts}
    }
    \caption{We conduct experiments on the impact of the layer of equipping the APG (a) and the impact of the number of generated prompts (b). All experiments are conducted on ImageNet-SubSet at non-pretrained 10-task incremental learning setting.
    }
            \label{fig:ab_hyper_params}
\end{figure}

\vspace{3pt}
\noindent\textbf{The impact of the location of the APG.}
As stated in \cref{sec:method}, the proposed APG takes the intermediate feature from layer $L_l$ and generates the class-specific prompts. 
In order to investigate the impact of the location $l$, we conduct experiments on $l=\{6,7,8,9,10\}$, and results are shown in \cref{fig:layer}.
It is shown that our method performs consistently when the APG is inserted to layers $l \in \{6,7,8,9\}$ but drops at layer $l = 10$.
It is mainly because the deeper level of the backbone encodes more task-specific information and it is not suitable for the APG for aggregating knowledge.
Our model performs best when $l=9$, and we choose $l=9$ as our default setting to balance the complexity and the performance. 

\vspace{3pt}
\noindent\textbf{The impact of the number of prompts.} 
As stated in \cref{sec:method}, the Multi-head Self Attention (MSA) is referred to perform the cross-attention operation.
Such a multi-head property allows us to generate any number of prompts to insert into deep layers.
To explore the impact of the number of prompts, we conduct experiments on $N_P \in \{0,1,2,3,4\}$, whose results are in \cref{fig:num_prompts}.
Compared with the baseline ($N_P=0$), we achieve 4.1\% accuracy improvement when generating 1 prompt.
When the number of prompts increases, the performance stays around 75\%.
This indicates that with the trainable APG, the generated prompt is powerful enough for a single prompt to instruct deeper layers.
Therefore, we choose $N_P=1$ as the default. 

\section{Conclusion}
In this work, we reveal the fact that existing prompt-based models achieve impressive performance in class-incremental learning but unfortunately they are affected by strong pretrained backbones.
The potential gap between the pretraining task and future tasks could be a stumbling block to existing prompting schemes in incremental learning.
We therefore contribute a learnable adaptive-prompting scheme.
The extensive experiments show that our method obtains satisfactory performance without pretraining and also achieves comparable performance to other models under strong pretraining.
In addition to adopting the knowledge from the pretraining task to other tasks, our work also shows that the prompt-based model can adaptively learn new knowledge when current knowledge is not sufficient for future tasks, which makes prompt-based models possible for more general incremental learning (e.g, more complicated tasks, multi-modality data).
\section{Acknowledgment}
This work was supported partially by the NSFC (U21A20471, U1911401, U1811461), Guangdong NSF Project (No. 2023B1515040025, 2020B1515120085).
\\
The authors would like to thank Kun-Yu Lin for insightful discussions.
The authors also thank three anonymous reviewers for their constructive suggestions.
{\small
\bibliographystyle{ieee_fullname}
\bibliography{egbib}
}

\end{document}